\documentclass[sigconf]{acmart}

\usepackage{adjustbox} 
\usepackage{verbatim}
\usepackage{enumitem}
\usepackage[T1]{fontenc}
\usepackage{graphicx}
\usepackage{subfig}
\usepackage{enumitem}
\usepackage{graphicx}
\usepackage{array}
\usepackage{multirow}
\usepackage{ragged2e}
\usepackage{multicol}
\usepackage{amsmath} 
\usepackage{color, colortbl}
\usepackage{hyperref}
\usepackage{tikz}
\usepackage{amsmath}
\usepackage{colortbl}
\usepackage{filecontents}

\usepackage{multirow}
\usepackage{tabularx}
\usepackage[linesnumbered,ruled,vlined]{algorithm2e}
\usepackage{makecell}
\usepackage{booktabs}
\usepackage{tikz}
\usepackage{amsmath}
\usepackage{textcomp}
\usepackage{bbding}
\usepackage{wasysym}

\usepackage{amssymb}
\usepackage{comment}
\usepackage{makecell}
\usepackage{mfirstuc}
\usepackage{graphicx}
\usepackage{xurl}
\usepackage{pifont}

\usepackage{listings}
\usepackage{minted}

\usepackage[utf8]{inputenc}
\usepackage{algorithm2e}

\usepackage{wrapfig}

\usepackage{url}
\usepackage{hyperref}
\usepackage{placeins}
\usepackage[normalem]{ulem}

\newcommand{\eg}{{\em e.g.,}\xspace}
\newcommand{\ie}{{\em i.e.,}\xspace}

\newcommand{\PP}[1]{\vspace{2px}\noindent{\bf#1.}}

\newcommand*\WC[1]{%
	\begin{tikzpicture}[baseline=(C.base)]
		\node[draw,circle,inner sep=0.2pt](C) {#1};
\end{tikzpicture}}

\newcommand*\BC[1]{%
	\begin{tikzpicture}[baseline=(C.base)]
		\node[draw,circle,fill=black,inner sep=0.2pt](C) {\textcolor{white}{#1}};
\end{tikzpicture}}

\newcommand{\cc}[1]{\mbox{\smaller[0.5]\texttt{#1}}}

\newcommand{\UP}[1]{\textcolor{black}{#1}} %

\newcommand{\sys}{\mbox{UnPII}\xspace}

\copyrightyear{2026}
\acmYear{2026}
\setcopyright{cc}
\setcctype{by}
\acmConference[ICSE-SEIP '26]{2026 IEEE/ACM 48th International Conference on Software Engineering}{April 12--18, 2026}{Rio de Janeiro, Brazil}
\acmBooktitle{2026 IEEE/ACM 48th International Conference on Software Engineering (ICSE-SEIP '26), April 12--18, 2026, Rio de Janeiro, Brazil}
\acmPrice{}
\acmDOI{10.1145/3786583.3786860}
\acmISBN{979-8-4007-2426-8/2026/04}

\begin{document}
\title{UnPII: Unlearning Personally Identifiable Information with Quantifiable Exposure Risk}


\author{Intae Jeon}
\email{intae515@g.skku.edu}
\affiliation{%
  \institution{Samsung Research}
  \city{Seoul}
  \country{South Korea}
}

\author{Yujeong Kwon}
\email{shr2008@g.skku.edu}
\affiliation{%
  \institution{Sungkyunkwan University}
  \city{Suwon}
  \country{South Korea}}

\author{Hyungjoon Koo}

\email{kevin.koo@skku.edu}
\authornote{Corresponding author.}
\affiliation{%
  \institution{Sungkyunkwan University}
  \city{Suwon}
  \country{South Korea}}

\begin{abstract}

The ever-increasing adoption of 
Large Language Models 
in critical sectors like 
finance, healthcare, and government 
raises privacy concerns regarding the
handling of sensitive 
Personally Identifiable Information (PII) during training.  
In response, regulations such as 
European Union's General 
Data Protection Regulation (GDPR) 
mandate the deletion of PII upon requests, 
underscoring the need for reliable
and cost-effective data removal solutions.
Machine unlearning has emerged as 
a promising direction for selectively 
forgetting data points.
However, existing unlearning techniques 
typically apply a uniform forgetting 
strategy that neither accounts for 
the varying privacy risks 
posed by different PII attributes 
nor reflects associated business risks.
In this work, we propose \sys, the first 
PII-centric unlearning approach that 
prioritizes forgetting 
based on the risk of individual or combined PII attributes. 
To this end, we introduce the 
PII risk index (PRI), 
a composite metric that incorporates multiple 
dimensions of risk factors: identifiability, sensitivity, usability, linkability, 
permanency, exposability, and compliancy. 
The PRI enables a nuanced evaluation of
privacy risks associated 
with PII exposures and
can be tailored to
align with organizational privacy policies. 
To support realistic assessment, we 
systematically construct 
a synthetic PII dataset 
(\eg 1,700 PII instances) that simulates 
realistic exposure scenarios.
\sys seamlessly integrates with 
established unlearning algorithms,
such as Gradient Ascent, 
Negative Preference Optimization, and 
Direct Preference Optimization,
without modifying their underlying principles. 
Our experimental results demonstrate 
that \sys achieves the improvements of 
accuracy up to $11.8\%$, 
utility up to $6.3\%$, and 
generalizability up to $12.4\%$,
respectively,
while incurring a modest fine-tuning overhead 
of $27.5\%$ on average during unlearning.

\end{abstract}

\maketitle             
\section{Introduction}

Today, the widespread adoption of Large Language Models (LLMs) 
is common in various sectors across domains such as 
finance~\cite{idbenjra2024investigating} 
(\eg credit scoring), 
healthcare~\cite{zhou2023survey,hu2021detection} 
(\eg diagnosis assistance, drug detection), 
government~\cite{sarzaeim2024framework,safaei2024end} 
(\eg public safety, policy analytics), and 
education~\cite{pal2024autotutor,li2023evaluating} 
(\eg tutoring support, content recommendation).
Those models are often trained on 
sensitive user data, including 
personally identifiable information 
(hereinafter referred to as PII), 
which can be unintentionally memorized 
and revealed during inference.

While direct and high-profile data breach,
such as Marriott~\cite{marriott2020}
(disclosing contact and 
payment information 
of 5.2 million individuals)
and Equifax~\cite{equifax2019}
(compromising records of 
over 140 million U.S. consumers) cases,
draw significant attention, 
the indirect exposures~\cite{mireshghallah-etal-2022-quantifying} 
from trained models
also raise privacy concerns.
In response, regulations such as 
the General Data Protection Regulation 
(GDPR)~\cite{regulation2018general}, 
the California Consumer Privacy Act 
(CCPA)~\cite{ccpa2018}, and 
China’s Personal Information Protection Law 
(PIPL)~\cite{pipl2021} mandate 
the complete elimination 
(\eg right to erasure or 
right to be forgotten) of PII upon 
user request or following 
a security breach.

To address such concerns, \emph{machine unlearning} 
has emerged as a promising direction 
for selectively removing specific
training data points without full model retraining.
In practice, however, efficient post-hoc 
data deletion in large-scale models 
presents several challenges.
First, removing individual records from models
with billions of parameters 
is computationally demanding in
the absence of original training 
corpus~\cite{bourtoule2021machine}. 
Second, unlearning strategies often
impair global model utility,
as the removal process inadvertently updates
knowledge beyond the intended deletion 
scope~\cite{maini2024tofu}. 
Third, unlearning must ideally 
support incremental and online operation 
to handle deletion requests as they arise, 
without disrupting service 
availability~\cite{izzo2021approximate}
while preserving acceptable
resource efficiency~\cite{nguyen2022survey}. 
Lastly, given the evolving landscape of 
global privacy regulations, the adaptability of unlearning 
techniques to shifting legal requirements 
constitutes a key operational advantage.
Collectively, these technical, 
operational, and 
regulatory hurdles highlight the need for 
specialized methodologies 
that enable efficient, reliable, 
and legally compliant 
data removal in production-scale models.

Early methods, such as
SISA~\cite{bourtoule2021machine},
improve efficiency 
by partitioning the dataset into 
isolated shards and retraining 
only those affected by deletion requests. 
However, such means face 
scalability challenges 
when applied to large-scale models (\eg LLMs)
due to their high computational cost. 
Gradient-based approaches~\cite{maini2024tofu,shi2024muse,bu2024unlearning}
update model parameters to reverse 
the influence of unlearning samples.
Meanwhile, preference optimization 
approaches~\cite{zhang2024negative,rafailov2023direct,maini2024tofu,mekala2024alternate} modify 
model outputs by penalizing 
undesirable responses and 
promoting preferred alternatives.
Recent advancements in 
parameter-efficient techniques~\cite{houlsby2019parameter,chen2023unlearn} reduce computational overhead by limiting 
updates to a small subset of parameters;
however, they often assume the availability of
reference models or a retention dataset 
to maintain original performance.
Another direction is entity-level unlearning, 
such as Opt-Out~\cite{choi2024optout}, 
which enables the removal of
entire entity embeddings 
(\eg users, items, records).

While the previous approaches are 
effective in certain unlearning contexts, 
their effectiveness may diminish 
when applied to PII-level forgetting 
on real-world datasets.
The limitation arises from two key challenges: \WC{1}
applying the same forgetting strategy to 
every PII attribute (\eg name, social security number) 
in the unlearning dataset may not be effective
without accounting for differences in 
privacy sensitivity (or risk), and \WC{2}
prior works often rely on a retention set,
which may be unavailable in 
(PII-relevant) practical scenarios.

In this work, we propose \sys, 
the first PII-centric unlearning approach 
that dynamically prioritizes PII forgetting
based on the privacy risk of 
individual or combined PII attributes. 
By design, \sys can be seamlessly incorporated with \emph{any
gradient-based} unlearning methods, 
such as Gradient Ascent (GA)~\cite{maini2024tofu}, 
Negative Preference Optimization (NPO)~\cite{zhang2024negative}, 
and Direct Preference Optimization (DPO)~\cite{rafailov2023direct}.
In essence, given a PII-containing model,
\sys begins with identifying PII in the model's output
through consultations with a large language model 
using PII-inducing queries. 
Next, \sys computes a PII risk index (PRI; ranging from 0 to 1), quantifying the exposure 
risk of PII, either individually or in combination.
Lastly, \sys unlearns the target dataset 
by applying a gradient-scaling loss function,
adjusting the forgetting signal based on 
the computed risk value.
To facilitate this, we construct a synthetic PII
dataset (\eg 1,700 PII examples), generating
a PII-containing model.
Besides, we introduce a quantifiable PII
risk assessment metric that evaluates
the privacy risk associated with 
the PII attribute exposure, which 
captures varying factors such as
identifiability, sensitivity, usability, 
linkability, permanency, exposability, 
and compliancy.

\UP{From an industrial MLOps (Machine 
Learning Operations) perspective, 
\sys can be seamlessly integrated into 
Continuous Training (CT) pipelines. 
Unlike full retraining that incurs 
prohibitive GPU costs and poses risks to 
service availability, \sys operates 
as a parameter-efficient fine-tuning 
module. 
This design allows practitioners to batch 
PII deletion requests and apply updates 
within standard CI/CD (Continuous 
Integration and Continuous Delivery) 
cycles, thereby reconciling strict privacy 
compliance requirements with operational 
efficiency.
}

Our empirical evaluation, conducted with
three baseline approaches 
(\eg GA, NPO, DPO)
across various forgetting ratios 
(\eg 1\%, 5\%, 10\%),
demonstrates that \sys outperforms these baselines, 
improving the harmonic mean 
(up to 5\%)
of accuracy (up to 11.8\%), 
utility (up to 6.3\%), and 
generalizability (up to 12.4\%), 
with a modest 
fine-tuning overhead (27.5\% on
average). 

The main contributions of our 
paper are as follows:
\begin{itemize}[label=\textbullet, leftmargin=*]
    \item
    We introduce \sys, the first PII-centric machine unlearning approach 
    that dynamically prioritizes the PII forgetting based on
    the risk of individual or 
    combined PII attributes.
    \item 
    We propose a quantifiable PII risk assessment metric 
    that evaluates the privacy risk associated with PII attribute exposure.
    \item 
    We construct a synthetic PII dataset (\eg 1,700 PII instances) 
    that simulates realistic exposure scenarios, providing
    a benchmark for evaluating the effectiveness of 
    PII unlearning techniques.
    \item 
    We integrate \sys with three (popular) 
    unlearning techniques (\ie GA, NPO, and DPO)
    and evaluate them in terms of
    accuracy, utility, and generalizability.
\end{itemize}

 We released our source code and 
    dataset to foster further research 
    in the field of machine unlearning 
    for privacy protection~\footnote{\url{https://github.com/SecAI-Lab/unpii}}.

\section{Background}

\PP{Machine Unlearning and Challenges}
Contemporary regulations such 
as GDPR (EU)~\cite{regulation2018general} 
and CCPA (California)~\cite{ccpa2018} enforce 
the \emph{right to be forgotten}, 
granting individuals the authority to 
request the deletion of their 
(potentially sensitive) personal data. 
In response, machine unlearning has 
emerged as a promising approach to 
forget a subset of 
data samples from a trained model.
Namely, given an initial model $f(\theta_{init})$ 
on the full dataset $D$, it aims to derive
an unlearned model $f(\theta_u)$ that
forgets the unlearning dataset ($D_f \subset D$),
while preserving the behavior of 
a reference model $f(\theta_r)$ 
retrained from scratch on the 
retention dataset ($D_r = D \setminus D_f$).

Formally, this objective can be written as:
\begin{equation}
\text{Unlearn}(f(\theta_{init}), D_f) = f(\theta_u) \approx f(\theta_r)
\end{equation}
However, machine unlearning faces 
several challenges. 
First, achieving selective forgetting is difficult
as many existing methods~\cite{bourtoule2021machine,kadhe2023fairsisa,maini2024tofu,zhang2024negative,rafailov2023direct,mekala2024alternate,chen2023unlearn,hu2024separate} treat 
all training data uniformly, limiting their 
ability to effectively remove specific 
sensitive or harmful samples. 
Second, unlearning introduces a trade-off between
forgetting effectiveness and overall model utility:
\ie aggressively forgetting targets may bring about
catastrophic forgetting, degrading performance 
on the retention dataset. 
Third, validating that the model has 
\emph{completely forgotten}
the designated samples remains open.
This paper focuses on PII-centric data samples,
integrating their associated risks into 
a quantifiable index during the unlearning process. 
Besides, we propose a risk assessment
metric to strike a balance among unlearning
accuracy, model utility, and generalizability.

\PP{PII Risk Assessment}
PII refers to any data that can directly or 
indirectly identify an individual, 
such as names, addresses, phone numbers, 
social security numbers, and passport numbers. 
This type of information is inherently 
sensitive, and its exposure can lead to 
privacy breaches, legal liabilities, 
and ethical concerns~\cite{staab2023beyond,kim2023propile,shao2023quantifying}.
To assess the risks 
associated with PII exposures, 
authoritative institutions such as 
the National Institute of Standards and Technology
(NIST)~\cite{mccallister2010guide}, 
the Department of Homeland Security 
(DHS)~\cite{department2017handbook}, and 
the Health Insurance Portability and Accountability Act 
(HIPAA)~\cite{act1996health}, 
have developed risk-based frameworks 
grounded in quantitative evaluation criteria. 
While each institution defines and 
evaluates PII risk within its 
respective domain, this leads 
to inconsistencies 
in risk assessments across contexts 
and organizations. 
This paper proposes a 
unified and quantifiable PII risk 
assessment metric, which
can integrate various criteria 
from these domain-specific approaches.

\begin{table}[t!]
\centering
\caption{
PII categories and representative 
PII attribute examples of sensitive information.
Disclosure of such private information can 
infringe on a person's privacy.
}
\resizebox{0.48\textwidth}{!}{%

\renewcommand{\arraystretch}{1.0}

\renewcommand{\arraystretch}{1.0}

\begin{tabular}{ll}
\toprule
\textbf{Category} & \textbf{Representative 
 PII attribute} \\
\midrule
\textbf{Basic} & \makecell[l]{Name, date of birth, gender, nationality} \\ \hline
\textbf{Contact} & \makecell[l]{Region address, detailed address, \\ work address,  email address, Phone number, \\ social media, personal website, blog} \\ \hline
\textbf{Identifiers} & \makecell[l]{Social security number, work permit number, \\ passport number,  Driver license number} \\ \hline
\textbf{Financial} & \makecell[l]{Bank account number, credit card number, \\ 
card expiration date, income information,\\ 
Card security code, Credit score, loan details, \\ 
tax records,  cryptocurrency wallet address} \\ \hline
\textbf{Biometric} & \makecell[l]{Fingerprint data, DNA information, 
\\ iris scan data, facial recognition data, \\ 
Voice recognition data} \\
\hline
\textbf{Medical} & \makecell[l]{Medical record, health insurance ID number, \\
Hospitalization record, disability status, \\ diagnosis history, mental health record} \\
\hline
\textbf{Employment-related} & \makecell[l]{Job title, employment history, salary, \\ Employee ID number} \\ \hline
\textbf{Education-related} & \makecell[l]{Student ID number, transcript} \\ \hline
\textbf{Digital Footprints} & \makecell[l]{IP/MAC address, device identifier, \\ browsing/search history} \\ \hline
\textbf{Location} & \makecell[l]{ZIP code, Vehicle registration number, \\ 
 real-time location, GPS coordinate} \\ \hline
\textbf{Legal} & \makecell[l]{Criminal record, bankruptcy filing, \\ 
driving record, court record} \\ \hline
\textbf{Miscellaneous} & \makecell[l]{Insurance policy number, E-signature, \\  call log, voice-mail data} \\
\bottomrule
\end{tabular}

}
\label{table:total_pii}
\end{table}

\section{PII Attributes and Risk Factors}
\label{sec:data_generation}

\begin{table}[t!]
\centering
\caption{
Seven PII risk factors for our quantitative 
assessment of PII leakage. 
Each factor captures a distinct 
facet of risk, 
including identifiability, sensitivity, 
usability, linkability, 
permanency, exposability, and compliancy.
Every factor of 
a PII attribute represents
a value in the range (0,1), which
is parameterized based on 
organizational policies or requirements.
}
\resizebox{0.99\linewidth}{!}{
    
\begin{tabular}{ll}
\toprule
\textbf{PII Risk Factor} & \textbf{Description} \\
\midrule
\textbf{Identifiability} & Uniqueness of the PII that can identify an individual \\
\textbf{Sensitivity} & Potential 
psychological and social harm upon the PII exposures \\
\textbf{Usability} & Usefulness of the PII for attackers in carrying out malicious actions\\
\textbf{Linkability} & Likelihood that the PII can be linked to other data sources \\
\textbf{Permanency} & Difficulty in changing or revoking the PII once exposed \\
\textbf{Exposability} & Frequentness or broadness
of the PII exposures during normal use \\
\textbf{Compliancy} & Severity of legal or regulatory consequences upon the PII breach \\
\bottomrule
\end{tabular}

}
\label{tbl:attr_des}
\end{table}

\PP{PII Categories and Attributes}
We analyze major 
PII breach incidents~\cite{McCandless2025}
that are publicly available, 
classifying PII attributes 
into 12 categories.
We confirm that severe privacy risks often arise from combinations of PII. 
For instance, the Marriott breach leaked contact details 
(\eg phone numbers, email, addresses) alongside 
personal attributes (\eg company, gender, birth date)~\cite{marriott2020}.
The Facebook breach exposed user IDs and phone numbers 
linked with names and location data~\cite{facebook2020}.
Meanwhile, the Equifax breach revealed full names,
 birth dates, SSNs, and addresses~\cite{equifax2019}, 
demonstrating the compounded risk of aggregated identifiers.
\autoref{table:total_pii} organizes 
representative PII attributes
with their corresponding classes.

\PP{PII Risk Factors}
With a thorough analysis 
of risk assessment from NIST~\cite{mccallister2010guide}, 
DHS~\cite{department2017handbook}, and 
HIPAA~\cite{act1996health},
we carefully define seven risk factors
capable of capturing a distinct aspect 
of risk, including identifiability,
sensitivity, usability, linkability, 
permanency, exposability, 
and compliancy as shown in~\autoref{tbl:attr_des}.
\UP{For instance, NIST specifies 
``directly identifiable'' (\eg SSNs) and 
``linkable'' attributes (\eg ZIP codes),
which map to high- and low-risk values,
respectively.}
These factors represent flexible values,
ranging from $0$ to $1$, which
we parameterize based on institutional
policies or privacy requirements.
Unlike existing factor-based assessments,
we identify \UP{the risk-driven 
factors tied to
the exposure of each PII attribute}.
These risk factors are 
configurable within a range of $[0,1]$,
allowing organizations to tailor them
to align with their privacy policies,
which further guide the computation of
per-sample unlearning intensity
(\S\ref{sec:pii_unlearning_fw}).
This alignment ensures 
stronger deletion for high-risk PII while 
permitting lighter treatment of low-risk cases, 
resulting in a more cost-effective, auditable, 
and defensible compliance posture.

\section{\sys: Unlearning PII with 
PII Risk Index}


\begin{figure*}[t!]
\centering
\includegraphics[width=0.7\linewidth]{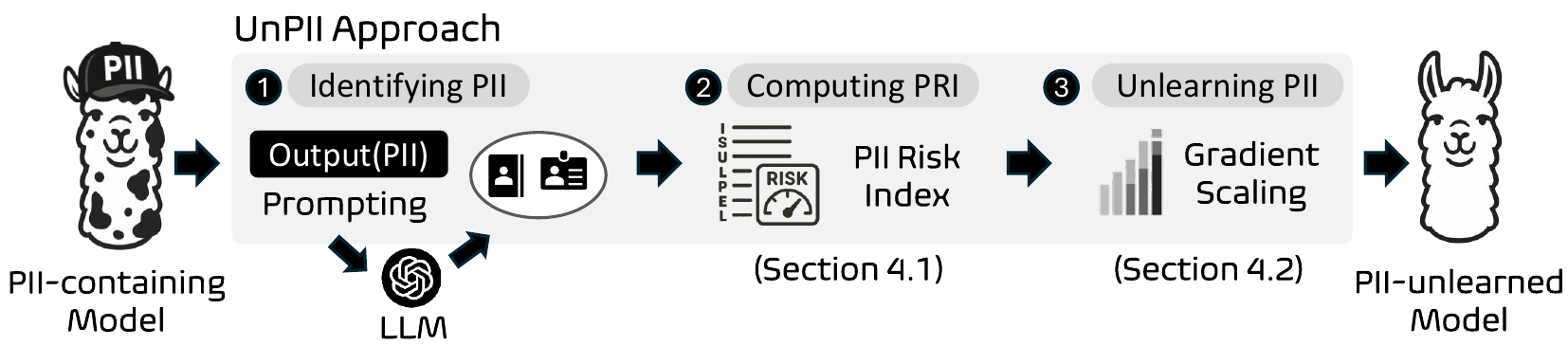}
\caption{
Overall \sys workflow for unlearning PII.
Given a model (\eg LLaMA2~\cite{touvron2023llama})
that produces outputs containing PII
(by PII-inducing questions),
\protect\BC{1} \sys identifies PII in the model's output by prompting 
an external LLM with a tailored query (\autoref{tab:identi_pii}).
\protect\BC{2} \sys then computes a PII risk 
index (PRI) that
quantifies the exposure risk,
either individually or in combination (\S\ref{sec:quantificfation_pii}).
\protect\BC{3} \sys integrates the index 
into the (existing) model's loss 
for unlearning via gradient scaling (\S\ref{sec:pii_unlearning_fw}),
generating a model that unlearns 
the target PII.
%
}

\label{fig:fw}
\end{figure*}

\PP{Approach Overview}
We assume a PII-containing model 
and an unlearning dataset in the absence
of a retention dataset.
We introduce \sys, 
\emph{a PII-centric unlearning approach that
is compatible with existing unlearning methods}.
As illustrated in~\autoref{fig:fw}, 
\sys operates in three stages. 
First, \sys identifies the PII
exposures by consulting an LLM, 
such as GPT-4o mini~\cite{liu2024evaluating},
to evaluate its output.
The rationale behind this approach builds
on a recent study~\cite{openai2023gpt4omini},
which systematically demonstrates that LLMs 
outperform traditional techniques,
such as regular expressions, keyword 
searches, and entity detection, in nearly all 
personal information extraction scenarios.
%
%
%
Second, \sys computes a quantifiable PII risk metric
for individual or combined PII attributes
(\S\ref{sec:quantificfation_pii}).
To exemplify, \autoref{table:pii_example} in Appendix 
displays $10$ individual and $7$ 
combined PII attributes,
assigning the values for seven risk factors.
Third, \sys unlearns the target dataset by
scaling the gradient (\ie incorporating the 
PII risk index into a loss function),
building a PII-unlearned model
(\S\ref{sec:pii_unlearning_fw}).
Notably, \sys is designed for seamless 
integration with existing approaches,
including gradient ascent, negative 
preference optimization, and direct 
preference optimization.

\subsection{Quantifiable Metric for PII Exposure Risk}
\label{sec:quantificfation_pii}
%
%


\PP{Metric Requirements}
Designing a quantifiable risk metric 
for PII is inherently challenging
due to the diverse aspects of 
individual PII attributes 
that affect their
susceptibility to leakage.
Besides, the perception 
and evaluation of PII attributes 
may vary across organizations 
or institutions.
In addition, combinations of PII 
can significantly amplify the risk
of re-identification; 
\eg the combination of 
zip code, birth date, and gender
can uniquely identify 87\% of the 
Americans~\cite{sweeney2000simple}.
Furthermore, the metric must provide 
interpretability in a quantitative form
to support objective assessment and 
comparison, rather than relying on 
subjective or qualitative evaluations
(\eg low, medium, high).

\PP{Metric Design}
We design a PII risk index (PRI)
to quantify the potential impact of 
exposing singular or aggregated
PII attributes, guided by the following principles:
the index should \WC{1} capture
multidimensional risk factors 
associated with each PII attribute; 
\WC{2} emphasize elevated risk 
when multiple PII attributes
are exposed together; and
\WC{3} express the overall risk
as a normalized value in the (straightforward)
range of (0, 1), where a higher value
represents a higher risk.
\UP{This design choice enables 
a flexible and practical interface 
for organizations by translating their 
internal privacy policies 
into quantitative values (\eg 
setting 0.9 for confidential data and 
0.1 for public data) through 
pre-defined intervals,
while preserving the underlying 
algorithm unchanged.}
%

\begin{figure}[!t]
\centering
\includegraphics[width=0.85\linewidth]{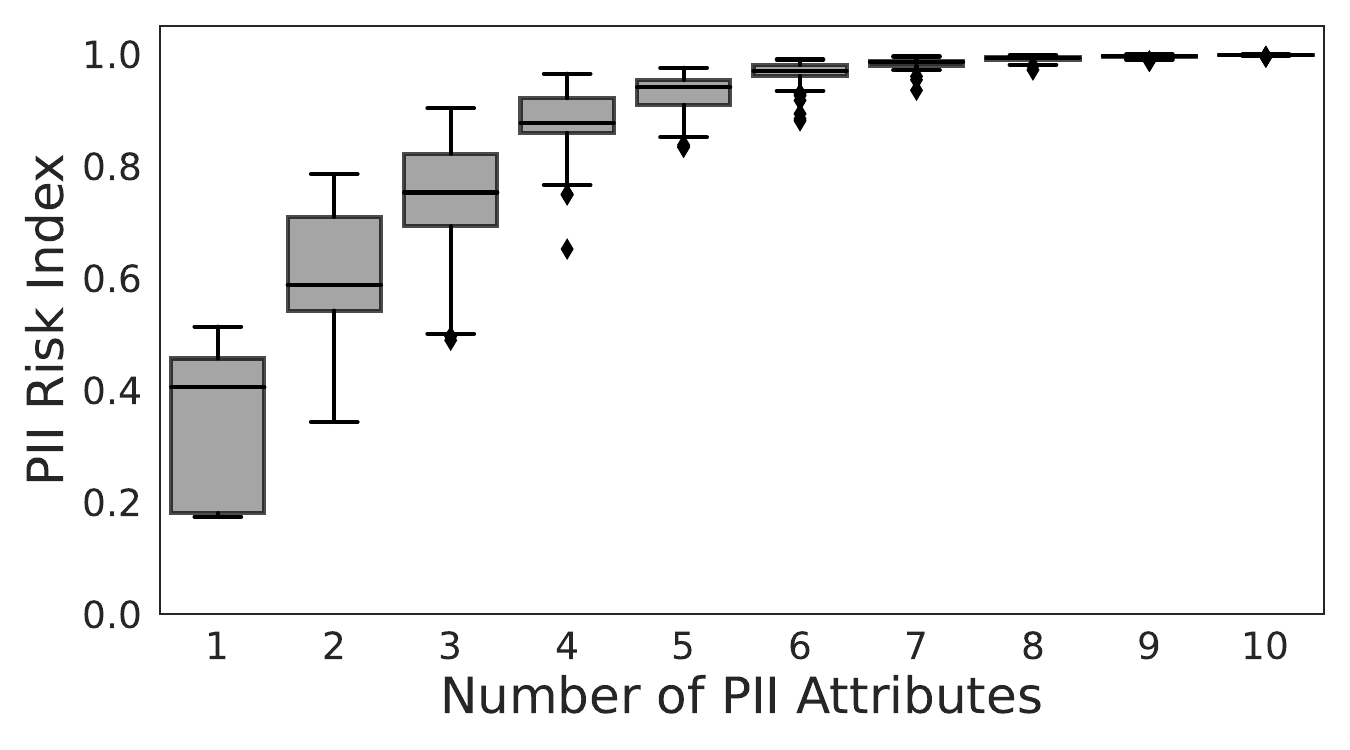}
\caption{
Distribution of PII risk indexes
(PRIs) across 1,000 simulations,
assuming the leakage of
one to ten PII attributes.
The results show that
as the number of exposed 
attributes increases,
the overall risk rises while the 
(standard) deviation decreases.
Five or more PII exposures
approach $1.0$ with a low standard deviation.
%
}
\label{fig:risk_boxplot}
\end{figure}

\PP{PII Risk Index for \sys}
%
%
%
We propose a \emph{PII risk index} 
or \emph{PRI metric}, which quantifies 
the risk associated with PII
upon disclosure, which
satisfies the aforementioned 
requirements.
%
%
Let $R$ denote PRI, which
incorporates $l$ exposed 
PII attributes, each
evaluated with 
$k$ distinct risk factors. 
For each attribute $i$ and
risk factor $j$, let $a_{ij} \in [0,1]$
denote the risk score, and
$w_{ij} \in [0,1]$ represent
the corresponding weight reflecting
the organization's policy preferences.
Each risk factor assesses
different dimensions of exposure risk,
such as the uniqueness, potential harm,
usefulness, linkability, breadth,
and legal consequences 
(\autoref{tbl:attr_des}).
Organizations may assign
zero weight to disregard
certain dimensions 
(\eg $w_{j}$ for usability).
The weights must be properly
normalized such that
$\sum_{j=1}^{k}w_{j}=1$.
Then, the individual risk $r$
is computed by aggregating 
the weighted risk scores 
via an inner product across 
risk factors and a summation 
across all attributes.
\UP{We parameterize the lambda term ($\lambda = 0.025$) 
to prevent premature saturation of the 
risk index towards 1.0 when the number 
of attributes is small, ensuring that 
gradient scaling remains sensitive to the 
addition of new risk factors.}
%
\begin{equation}
    r = \lambda k l + \sum_{i=1}^{l}\prod_{j=1}^{k} w_{ij}a_{ij}
    \label{eq:risk}
\end{equation}
The additive term ($\lambda kl$)
compensates for counterintuitive 
risk dilution as the number 
of attributes or factors increases.
Finally, we apply
the hyperbolic tangent function
to bound the resulting PRI value
within the open interval (0,1).
\begin{equation}
    R = \tanh(r) = \frac{e^{r}-e^{-r}}{e^{r}+e^{-r}} \in (0,1) \quad\text{where}\quad r > 0
    \label{eq:risk}
\end{equation}
\autoref{fig:risk_boxplot} 
illustrates the 
distribution of PRI values 
across $1,000$ simulations
with~\autoref{table:pii_example} 
in Appendix,
assuming the leakage of one to ten PII attributes. 
Notably, PRI nears $1.0$ 
under exposures of five or more PII attributes, 
while maintaining a low standard deviation.

\subsection{\sys: Unlearning PII with its Risk Index}
\label{sec:pii_unlearning_fw}
By design, \sys can be seamlessly 
integrated with 
other machine unlearning techniques.
To enhance the efficiency of eliminating PII attributes, 
we adopt gradient scaling~\cite{micikevicius2017mixed}
that assists in stabilizing the selective gradient
updates when unlearning a small subset of data samples.
Indeed, different unlearning models incorporate
customized loss functions to control the forgetting rate:
\ie by amplifying the forgetting signal while
mitigating unintended side effects on retention data.
In this paper, we apply three 
optimization-based 
unlearning techniques: GA~\cite{maini2024tofu}, NPO~\cite{zhang2024negative}, and DPO~\cite{rafailov2023direct}.
In essence,
once PII attribute(s) are revealed,
\sys computes the corresponding PRIs
(\S\ref{sec:quantificfation_pii}),
followed by plugging them
into a tailored loss that can guide 
a model to unlearn them.

\PP{Loss Function in Gradient Ascent}
The primary goal is to derive the model $\pi_{\theta}$ for unlearning by updating the pre-trained model $\pi$ to forget PII, 
using the forget set $D_f = \{x_f, y_f\}$, 
where $x_f$ and $y_f$ represent the forget prompt and forget response, respectively.
The loss associated with $D_f$ is increased using gradient ascent, 
which prevents $\pi_{\theta}$ from generating $y_f$.
The loss function explicitly aims to reverse the training effects from $D_f$ in $\pi$.

 
\begin{equation}
\mathcal{L}_{\text{GA}} \ \dot{=}\ 
-\log \pi_\theta(y_f \mid x_f)
\label{eq:ga}
\end{equation}

\PP{Loss Function in Negative Preference Optimization}
NPO encourages $\pi_\theta$ to assign a lower probability to $y_f$, 
thereby learning an implicit dispreference for $y_f$. 
By using $\pi$, it reduces the prediction difference between the two models, 
helping to maintain the stability of $\pi_\theta$.
The outer sigmoid function $\sigma$ introduces a smooth, bounded transformation that stabilizes gradients and prevents exploding loss values during training.
Here, $\beta$ is a scaling factor that regulates the strength of the regularization.

\begin{equation}
    \mathcal{L}_{\text{NPO}} \ \dot{=}\ 
    -\frac{2}{\beta} \log \sigma\left( 
        - \beta \log \frac{\pi_\theta(y_f \mid x_f)}{\pi(y_f \mid x_f)} 
    \right)
\label{eq:npo}
\end{equation}

\PP{Loss Function in Direct Preference Optimization}
DPO~\cite{rafailov2023direct} compares a preferred response $y_p$ and $y_f$ given the same prompt $x_f$.
It encourages $\pi_{\theta}$ to assign a higher probability to $y_p$ than to $y_f$.
To ensure stable model learning, this formulation uses $\pi$, 
which helps maintain model stability.
The outer sigmoid function $\sigma$ smooths the loss landscape and bounds the gradients, which contributes to stable optimization.
The approach is designed to reverse the preference behavior learned from $D_f$, 
with $\beta$ regulating the strength of the regularization.

\begin{equation}
    \mathcal{L}_{\text{DPO}} \ \dot{=}\ 
    -\frac{2}{\beta} \log \sigma\left( 
        \beta \log \frac{\pi_\theta(y_p \mid x_f)}{\pi(y_p \mid x_f)} 
        - 
        \beta \log \frac{\pi_\theta(y_f \mid x_f)}{\pi(y_f \mid x_f)} 
    \right)
\label{eq:dpo}
\end{equation}


\PP{Gradient Scaling in \sys}
The gist of \sys lies in its 
\emph{gradient scaling mechanism}, 
which dynamically adjusts a base loss 
function, $\mathcal{L}_{\text{base}}$, 
according to the PII Risk Index ($R_p$). 
The final loss function is defined by
multiplying $\mathcal{L}_{\text{base}}$ 
with a PRI-based scaling factor as follows:
\begin{equation}
    \mathcal{L}_{\text{\sys}} = \mathcal{L}_{\text{base}}(1 + R_p)
    \ \text{where} \ 
    \mathcal{L}_{\text{base}} \in \{ 
        \mathcal{L}_{\text{GA}}, \mathcal{L}_{\text{NPO}}, \mathcal{L}_{\text{DPO}} 
    \}
\label{eq:unpii}
\end{equation}
Note that the risk index \(R_p\) is 
computed individually for each sample 
within an unlearning batch. 
Consequently, the scaling operates at the
per-sample level: each sample's loss 
is modulated by $(1+R_p)$,
and the final batch loss is obtained by 
aggregating these risk-weighted terms
(\eg via mean or sum). 
The scaling factor directly controls 
the \emph{strength of forgetting} 
in proportion to 
the quantified PII risk:
a higher-risk PII corresponds to 
a larger \(R_p\), thereby
amplifying the associated loss and
producing stronger gradient signals 
for unlearning. 
In our experiments, we instantiate 
\sys to three different base losses: 
\(L_{\mathrm{GA}}, 
L_{\mathrm{NPO}},\) and \(L_{\mathrm{DPO}}\).

\section{Evaluation}
\label{s:eval}
We evaluate \sys with varying 
experiments on a 64-bit 
Ubuntu 22.04 system with an 
AMD EPYC 7763 CPU @ 2.45GHz, 1TB RAM, 
and a single NVIDIA A100 GPU with 80GB of graphics memory.

\PP{Research Questions}
We formulate three research questions, 
each addressing a distinct aspect 
of the problem: effectiveness, 
consistency, and efficiency.

\begin{itemize}[label=\textbullet, leftmargin=*]
    \item
    (RQ1) To what extent does integrating \sys with existing 
    unlearning techniques enhance 
    overall performance
    (\ie Harmonic mean of accuracy, utility, and generalizability)?
     Besides, how does
     \sys affect the balance 
     among performance metrics
     under varying settings
     (\S\ref{sec:rq1})?
     \item 
    (RQ2) How consistently does \sys perform
    across different unlearning instances
    (\eg using random sampling)
    (\S\ref{sec:rq3})? 
    \item 
    (RQ3) How efficient is \sys 
    across different 
    unlearning approaches,
    including GA, DPO, and NPO
    (\S\ref{sec:rq4})?
 \end{itemize}
 
\subsection{Experimental Setup}
\label{ss:expsetup}

\PP{Dataset Construction}
We construct a synthetic PII dataset using 
GPT-4o~\cite{openai2024gpt4o}, as no existing
pseudo-PII dataset meets the needs of our study.
To prevent any association with real individuals, 
all prompts are carefully crafted to generate entirely 
artificial data.
We chose 10 representative PII attributes from each
category in~\autoref{table:total_pii},
along with 7 combinations of these attributes,
resulting in 1,700 samples (100 per attribute or combination).
While \sys allows for parameterizing 
risk dimensions 
according to institutional policy, 
for this experiment, we leverage 
GPT-4o~\cite{openai2024gpt4o} to assign values 
across seven risk dimensions, assuming 
its semantic understanding reflects real-world interpretations.
\autoref{table:pii_example} in Appendix
summarizes the assigned risk factor 
and the resulting PRI 
based on Equation~\ref{eq:risk}, 
using $k=7$ and $\lambda=0.025$.
As a final note, we define 
three datasets based on 
different forgetting ratios: \cc{forget01}, 
\cc{forget05}, and \cc{forget10} correspond
to 1\% (17 samples), 5\% (85 samples), 
and 10\% (170 samples) of 
the 1,700-sample dataset, respectively.
The prompts for data generation 
are displayed in Appendix~\autoref{tab:gen_prompt}.

\PP{Unlearning Validity}
In the absence of a universally 
accepted method for 
validating unlearning status, we adopt 
a two-step matching strategy 
tailored to the structural properties of
each PII attribute: \textit{pattern matching} 
and \textit{semantic matching}.
The first step applies pattern matching to detect
structured PII attributes such as phone numbers and
social security numbers.
We use pre-defined regular expressions
to identify such instances; if a model's output
matches the pattern and contains the correct value,
it is considered a failure to forget, otherwise a success.
The second step applies semantic matching 
for other unstructured PII attributes, 
including names, addresses, and hospital records.
We leverage a commercial large language model 
(\eg GPT-4o mini~\cite{openai2023gpt4omini}) 
to assess whether the model’s output 
contains any explicit or implicit PII-related instances. 
\autoref{tab:eval_prompt} in Appendix shows prompt examples for evaluation.

\PP{Evaluation Metrics for Unlearning}
Building on the unlearning validation 
approaches described above, 
we introduce three key metrics to assess 
unlearning performance: accuracy (A), 
utility (U), and generalizability (G).
First, accuracy quantifies the model's ability 
to effectively forget targeted PII attributes, 
evaluated on the unlearning dataset.
Second, utility measures the extent to which 
the model preserves performance 
on non-PII content after unlearning. 
Third, generalizability captures whether 
the model forgets PII in unseen instances
present during training but excluded
from the unlearning set. 
To provide a comprehensive evaluation 
of unlearning quality, we report H-AUG,
the harmonic mean of accuracy, utility, and 
generalization as the following equation.
\begin{equation}
\text{H-AUG} = \frac{3}{\frac{1}{A} + \frac{1}{U} + \frac{1}{G}} 
\label{eq:haug}
\end{equation}
This formulation promotes 
balanced unlearning by penalizing 
the overall score when any individual 
metric is low. 
For robustness, all experiments are 
conducted three times, and the reported 
results reflect the average performance.

\subsection{Effectiveness of \sys}
\label{sec:rq1}
\begin{figure}[!t]
\centering
\includegraphics[width=0.95\linewidth]{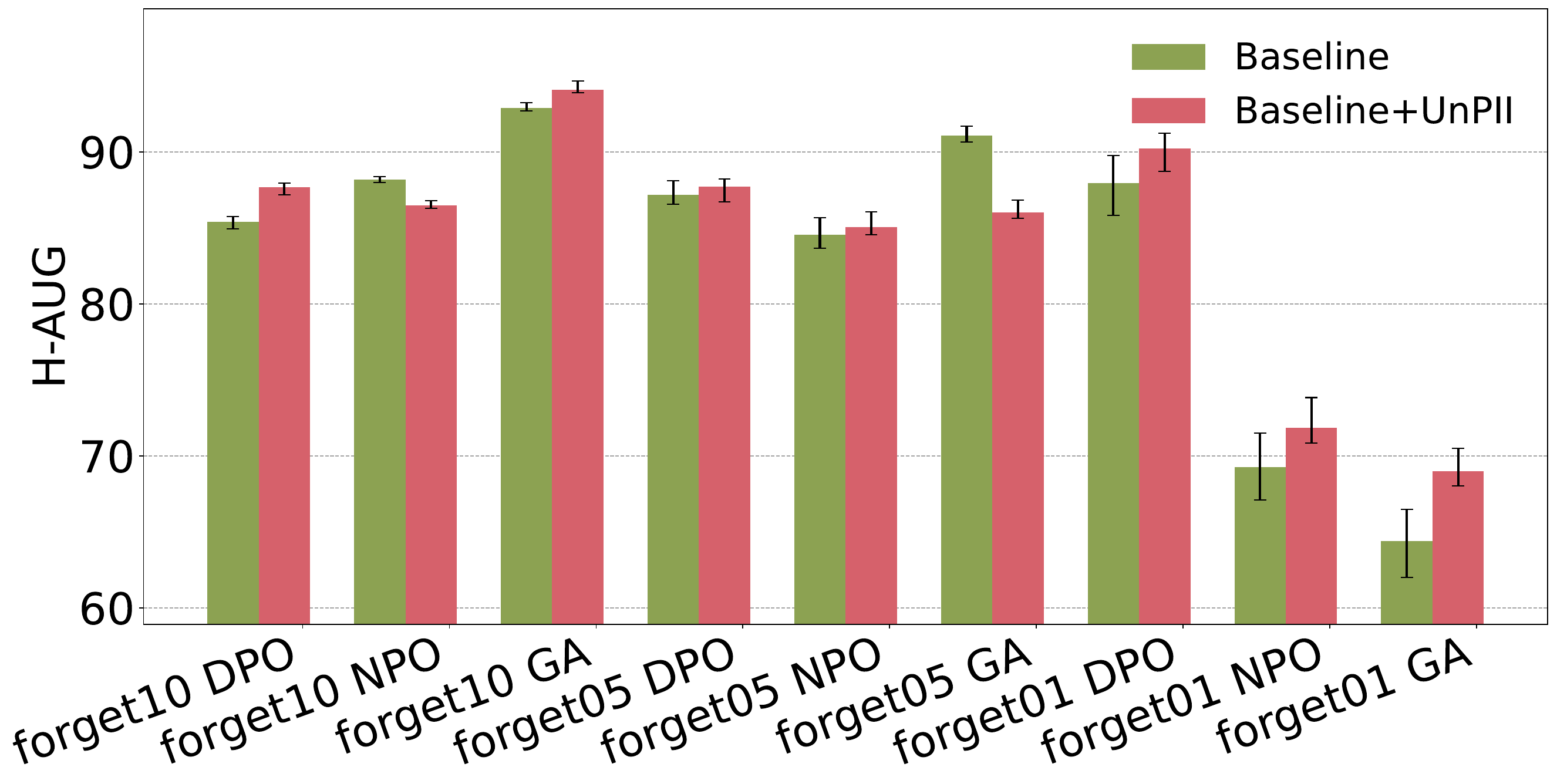}
\caption{
This figure presents a comparison of the performance of three unlearning methods (GA~\cite{maini2024tofu}, NPO~\cite{zhang2024negative}, and  DPO~\cite{rafailov2023direct}) with and without our \sys technique across three forgetting ratio settings (\cc{forget10}, \cc{forget05}, \cc{forget01}). 
Green lines indicate baseline results, while red lines indicate results with \sys. 
Each experiment was repeated three times, and black error bars indicate variation across runs.
}
\label{fig:rq0_bar}
\end{figure}

In this section, we evaluate the effectiveness of \sys 
when combined with GA~\cite{maini2024tofu}, 
DPO~\cite{rafailov2023direct}, 
or NPO~\cite{zhang2024negative} across
three unlearning datasets corresponding to
1\%, 5\%, and 10\% forgetting ratios.
As depicted in~\autoref{fig:rq0_bar}, \sys
yields overall improvements in H-AUG 
(up to $5\%$)
compared to the baseline unlearning approaches.

\label{sec:rq1}

\begin{figure}[!t]
\centering
\includegraphics[width=0.99\linewidth]{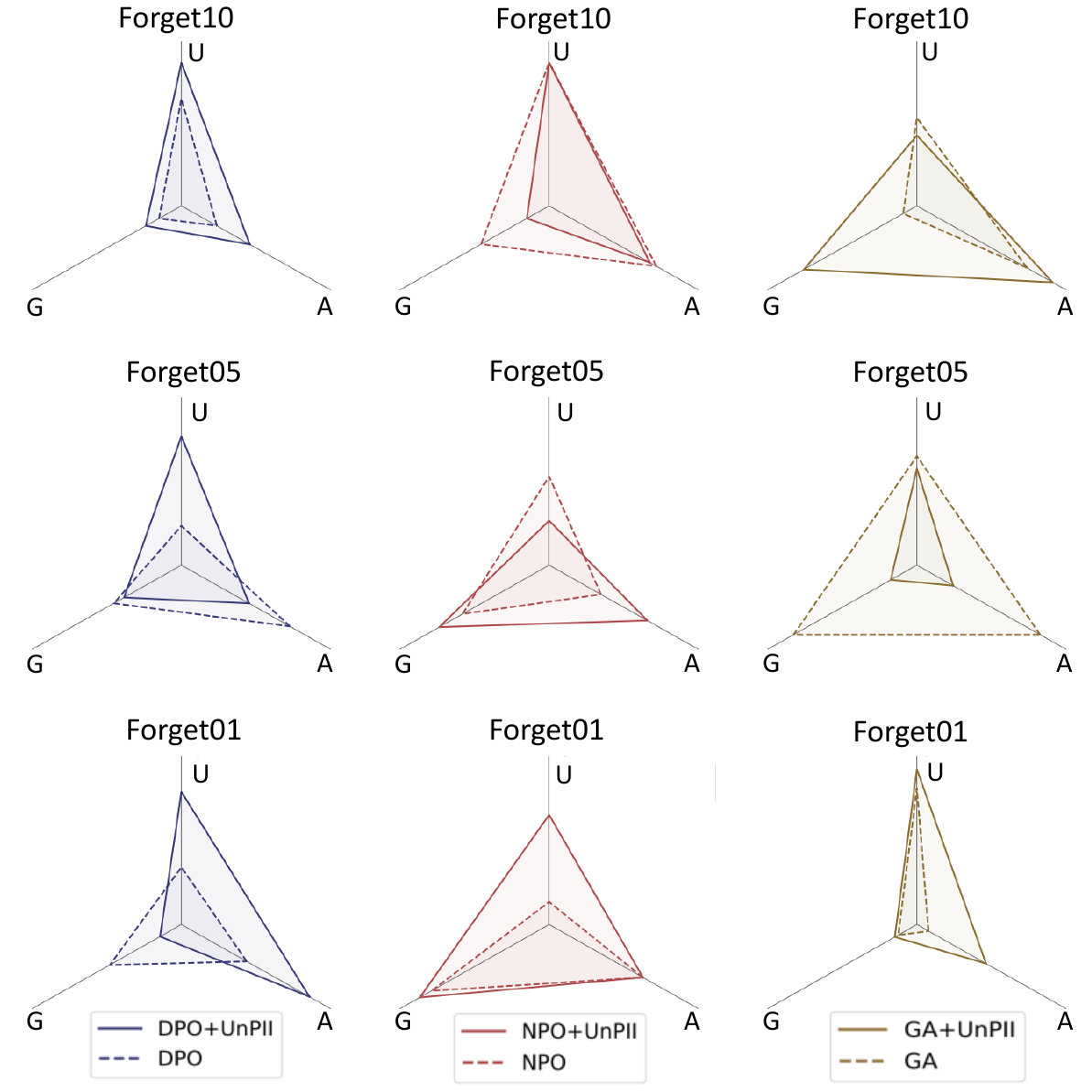}
\caption{
Performance breakdown
by accuracy (A), utility (U), 
and generalizability (G)
illustrating the comparative results 
(measured by the harmonic mean; H-AUG) 
of three unlearning techniques -- 
DPO~\cite{rafailov2023direct} \UP{(left)}, NPO~\cite{zhang2024negative} \UP{(middle)}, and GA~\cite{maini2024tofu} \UP{(right)} -- and
their variants incorporating \sys
under three forgetting ratio settings 
(\cc{forget10}, \cc{forget05}, \cc{forget01}).
Solid lines denote the performance with \sys, 
while dashed lines correspond to the baselines. 
Notably, larger solid triangle areas indicate that
\sys enhances overall performance across most configurations.
}
\label{fig:rq0_radar}
\end{figure}

\PP{Performance Enhancement Across Unlearning Approaches}
Approach-wise,
\autoref{fig:rq2_graph} highlights notable differences
in the effectiveness of each technique.
The DPO~\cite{rafailov2023direct} approach exhibits 
strong performance in the early steps
at the $10\%$ forgetting ratio, but shows a sharp decline
after a certain threshold; in contrast, 
DPO+\sys demonstrates a more gradual performance
degradation across intervals. 
The NPO~\cite{zhang2024negative} approach maintains
a relatively high average performance and 
remains stable throughout the mid-steps 
under the 5\% and 10\% forgetting ratios.
Yet, under the 1\% forgetting ratio, 
it experiences a quick performance drop in 
the later steps following an initial rise.
Similarly, this pattern is observed in NPO+\sys. 
Meanwhile, both GA~\cite{maini2024tofu} and GA+\sys 
display substantial early-step improvements 
under certain conditions. 

\begin{figure}[!t]
\centering
\includegraphics[width=1.1\linewidth]{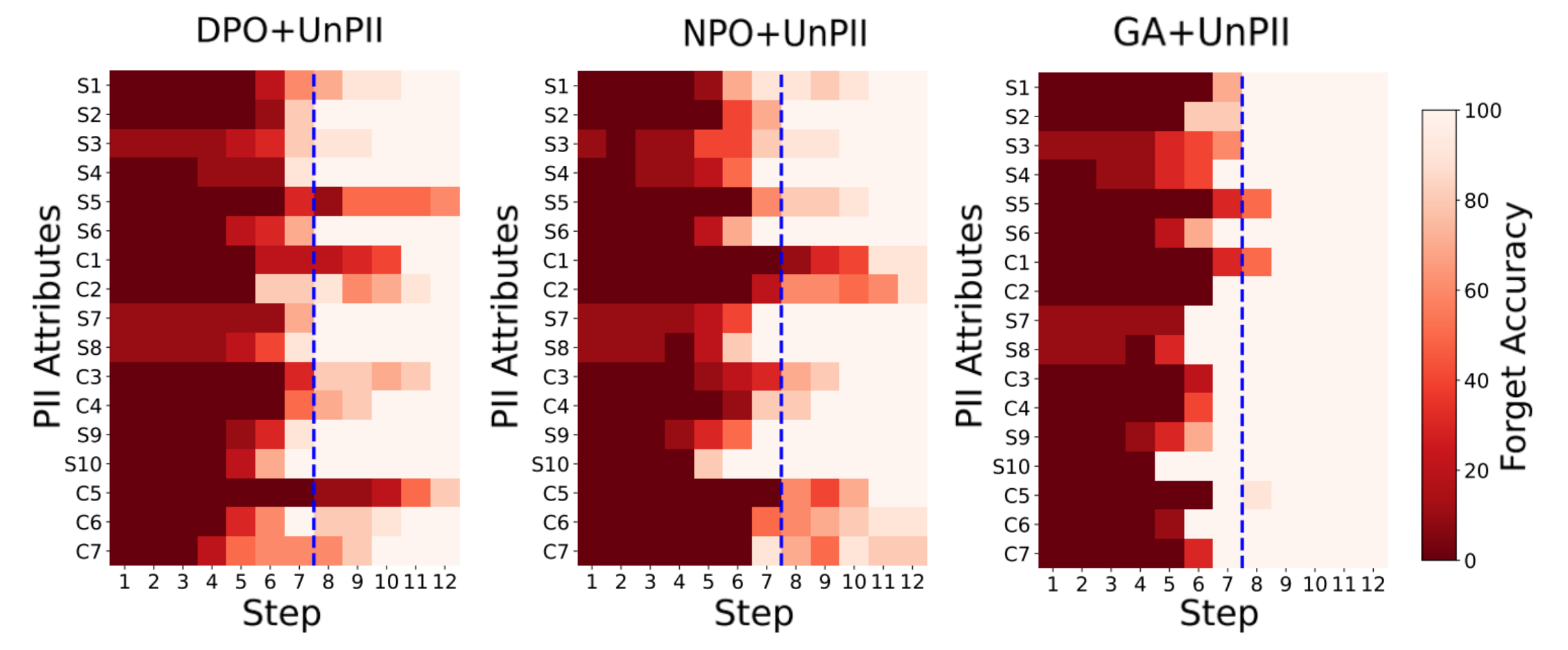}
\caption{
Comparison of forget accuracy
over training steps for
different \sys-applied  
unlearning techniques:
DPO+\sys, NPO+\sys, and GA+\sys.
Each heatmap represents the progress
of unlearning PII attributes
(order by PRI values),
with a transition from dark to light colors.
While demonstrating the effectiveness
of \sys with all unlearning approaches,
we observe that incorporating \sys with GA 
outperforms than others slightly.
}
\label{fig:rq4_heatmap}
\end{figure}

\PP{Decomposed Performance Analysis: Accuracy, Utility, and Generalizability}
We decompose H-AUG into 
individual performance metrics 
for all baseline methods.
\autoref{fig:rq0_radar} demonstrates 
that \sys achieves the improvements of 
accuracy up to $11.8\%$, 
utility up to $6.3\%$, and 
generalizability up to $12.4\%$,
respectively.
The forgetting accuracies
of models integrated with \sys 
mostly surpass the baselines
by around 5\% at each step,
demonstrating its effectiveness.
Likewise, incorporating \sys
positively impacts
model utilities: \eg
DPO+\sys and NPO+\sys exhibit
up to 8\% performance enhancements
over their baselines, 
while GA+\sys with a modest gain 
of around 2\%.
Furthermore, \sys contributes to
generalizability (evaluated on 
an unseen dataset), with
improvement of
up to approximately 3\%
compared to each baseline.
\autoref{fig:rq4_heatmap} presents step‑wise forget‑accuracy heatmaps for DPO+\sys, NPO+\sys, and GA+\sys, illustrating that GA+\sys converges slightly faster in unlearning high‑risk PII attributes. 
Notably, we observe certain configurations,
such as GA+\sys with the \cc{forget05} dataset,
deviate from overall trends, 
which we discuss in~\S\ref{s:discussion}.

\subsection{Consistency of \sys on Different Unlearning Samples}
\label{sec:rq3}
\begin{figure*}[!t]
\centering
\includegraphics[width=0.99\linewidth]{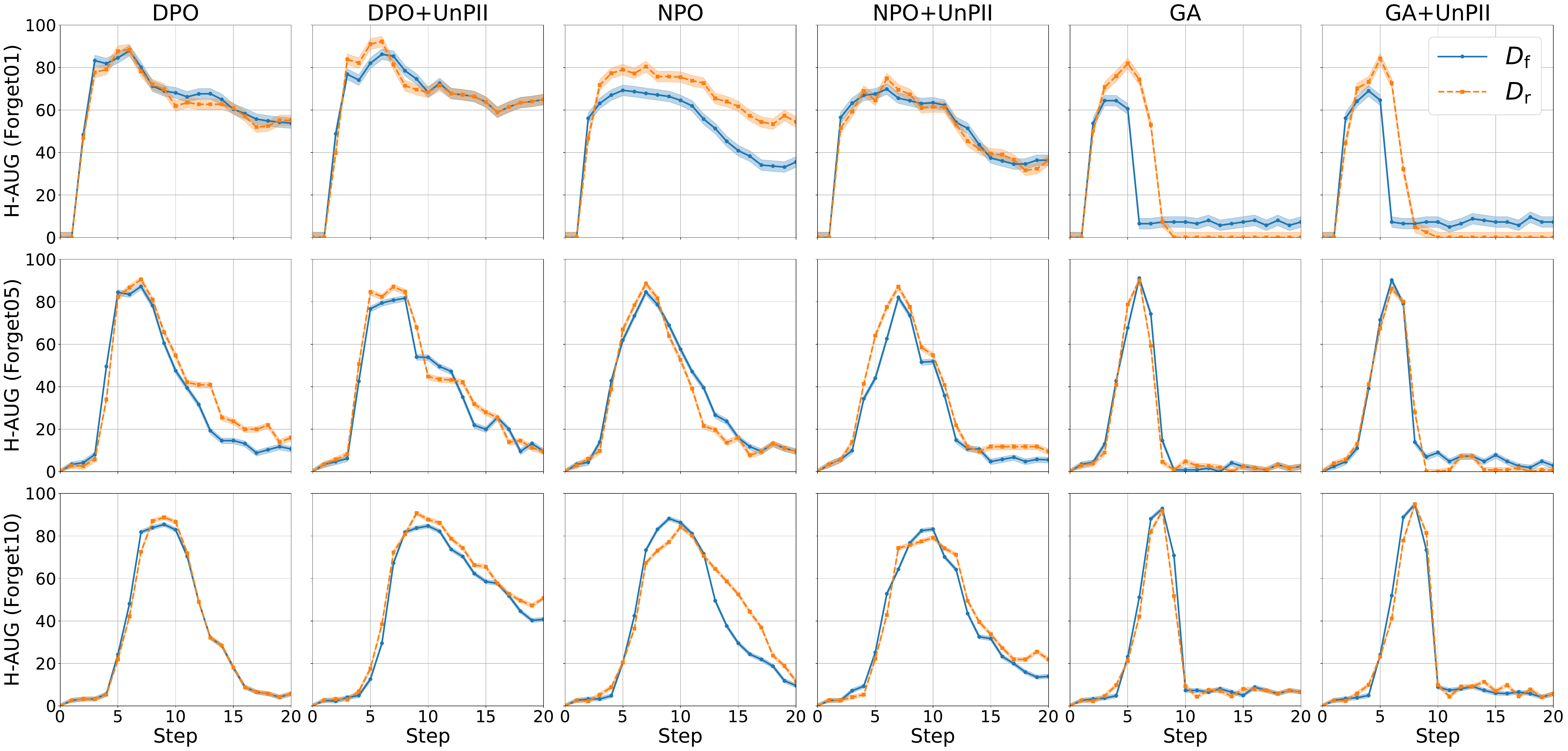}
\caption{
Comparison of unlearning performance
between the original unlearning dataset 
($D_f$) and a randomly re-sampled dataset ($D_r$) 
for GA~\cite{maini2024tofu}, 
NPO~\cite{zhang2024negative}, and 
DPO~\cite{rafailov2023direct}, with 
and without our \sys technique
across different forgetting ratios 
(\cc{forget10}, \cc{forget05}, \cc{forget01}).
Blue lines indicate 
results on $D_f$, 
and orange lines 
represent results on $D_r$. 
Despite marginal variations in the dataset 
composition, the overall performance 
remains consistent
(Section~\ref{sec:rq3}).
}
\label{fig:rq2_graph}
\end{figure*}
This section evaluates the performance consistency 
of the \sys approach when trained on
different unlearning samples
and integrated with existing unlearning methods:
\eg the original forgetting set ($D_f$) and 
a re-sampled forgetting set ($D_r$).
To this end, we compare the two models
trained on $D_f$ and $D_r$ across
varying forgetting ratios (\eg $1\%, 5\%, 10\%$).
The re-sampled sets are constructed to 
ensure that at least one PII attribute 
from each category is included.
This experiment aims to confirm whether the 
performance of the unlearned model (\sys) 
remains persistent, regardless of 
unlearning data samples.
As illustrated in \autoref{fig:rq2_graph},
H-AUG remains largely consistent
in most cases, indicating that 
\sys is robust to variations in the composition
of unlearning instances.

\subsection{Efficiency of \sys}
\label{sec:rq4}
We assess the efficiency of \sys
by measuring fine-tuning durations
over 20 unlearning steps across
different unlearning approaches.
As shown in~\autoref{tbl:exp_time},
the overall overhead averages $27.5\%$ with
increases of 21.6\% for GA (465.7s $\rightarrow$ 562.0s), 
25.3\% for DPO (628.1s $\rightarrow$ 790.1s), and 
35.6\% for NPO (474.4s $\rightarrow$ 642.3s).
However, this overhead is relatively modest
compared to retraining the entire model from scratch:
\eg the original LLaMA2 7B  required
$184,320$ GPU hours.
Detecting PII via the 
GPT‑4o‑mini~\cite{openai2023gpt4omini}  
interface (\ie API) causes dominant overheads; however,
it incurs an inexpensive cost of only \$0.01 
(\eg 320 API calls) across the entire run.
Notably, memory usage remains identical 
between baselines with and without \sys.

\begin{table*}[!t]
\small
\centering
\caption{
Comparison of training time overheads 
(in seconds) over 20 steps between 
the baselines and our 
PII-centric unlearning (\sys).
Overall,  GA+\sys exhibits the lowest overhead
of $\uparrow$ 21.6\% on average, while 
DPO+\sys and NPO+\sys record
25.3\% and 35.6\%, respectively.
Most of the time overheads arise from
inferences for identifying PII attributes
using GPT-4o mini~\cite{openai2023gpt4omini}.
}

\begin{tabular}{
    @{\hspace{6pt}}l
    @{\hspace{12pt}}c
    @{\hspace{18pt}}c
    @{\hspace{18pt}}c
    @{\hspace{18pt}}c
    @{\hspace{18pt}}c
    @{\hspace{18pt}}c
}
\toprule
 & \textbf{DPO} & \textbf{DPO+UnPII} & \textbf{NPO} & \textbf{NPO+UnPII} & \textbf{GA} & \textbf{GA+UnPII} \\
\midrule
\cc{forget01} & 686.7 & 889.0 ($\uparrow$ 29.5\%) & 558.0 & 738.3 ($\uparrow$ 32.4\%) & 550.3 & 617.7 ($\uparrow$ 12.3\%) \\
\cc{forget05} & 533.0 & 636.3 ($\uparrow$ 19.4\%) & 423.3 & 549.7 ($\uparrow$ 29.9\%) & 430.7 & 529.0 ($\uparrow$ 22.8\%) \\
\cc{forget10} & 664.7 & 845.0 ($\uparrow$ 27.1\%) & 442.0 & 639.0 ($\uparrow$ 44.6\%) & 416.0 & 539.3 ($\uparrow$ 29.6\%) \\
\midrule
Average & 628.1 & 790.1 ($\uparrow$ 25.3\%) & 474.4 & 642.3 ($\uparrow$ 35.6\%) & 465.7 & 562.0 ($\uparrow$ 21.6\%) \\
\bottomrule
\end{tabular}

\label{tbl:exp_time}
\end{table*}
 
\section{Implementation}

\PP{PII-containing Model for \sys}
We train the PII-containing model on 
our in-house synthetic dataset,
which include 1,700 cases that contain
17 PII attributes, using
the LLaMA2 7B~\cite{touvron2023llama} 
model implemented via the Hugging Face
Transformers library~\cite{wolf-etal-2020-transformers}. 
For parameter-efficient fine-tuning, 
we employed the LoRA~\cite{hu2022lora}
technique, which introduces 
two hyperparameters: 
$r$, the rank of the low-dimensional 
matrices to approximate weight 
updates, and $\alpha$, a scaling factor 
applied to these updates. 
We set $r=16$ and $\alpha=32$
in our experiments. 
The model has been optimized using 
AdamW with a learning rate 
of $2 \times 10^{-4}$ and
a weight decay of $0.01$, 
trained for $30$ epochs 
with a batch size of $16$.

\PP{PII-unlearned Model with 
Baseline Unlearning Approaches}
We train three PII-unlearned models
for \sys on the following three baseline
approaches: GA~\cite{maini2024tofu},
NPO~\cite{zhang2024negative} and 
DPO~\cite{rafailov2023direct}.
All models are trained under the 
same experimental settings as 
their respective baselines.
We train each model 
for 20 epochs with a batch size of 16, 
using the AdamW optimizer with a learning rate of $1 \times 10^{-5}$ 
and a weight decay of $0.01$.
In our experiments, each epoch corresponds to 
2, 6, and 11 steps for \cc{forget01},
\cc{forget05}, and \cc{forget10}, respectively.
For NPO and DPO, 
we set the scaling factor $\beta = 0.1$,
as used in the original implementations. 
Additionally, DPO is trained with 
the phrase ``I don’t know.'' specified
as the preferred response.

\section{Threats to Validity}
\label{s:discussion}

\PP{\UP{Limitations of Synthetic Dataset}}
The empirical evaluation primarily 
relies on (in-house) synthetic datasets 
to simulate controlled PII exposure 
scenarios. 
While an artificial dataset 
allows for systematic and 
reproducible analysis, it may 
not fully capture real-world complexities 
such as \UP{data noise and} sparsity,
the distribution of class imbalance, 
\UP{long-tail formats, or 
locale-specific identifiers.}
\UP{Additionally}, 
the process of generating 
synthetic samples using GPT-4o has 
inherent limitations, 
as it may have created 
attributes that are factually incorrect or 
inconsistent, such as an address with 
a mismatched city and ZIP code, 
or an identification number with 
a non-standard format. 
Hence, further validation on 
real-world datasets from 
diverse domains (\eg healthcare, finance) 
is essential to assess the practical 
applicability and robustness of
the proposed approach.

\PP{\UP{Limitations on Defining 
Risk Factors}}
\UP{While authoritative guidelines 
serve as our proxy for expertise,
the defined risk factors may not 
precisely correspond to 
real-world compliance standards across 
diverse domains and viewpoints,
as our work lacks direct legal review or validation from domain experts.}

\PP{Unlearning Assessment Metric}
As discussed in~\ref{ss:expsetup}, defining
a standard metric for evaluating 
whether a data point has been  
\emph{truly forgotten} through unlearning 
remains an open challenge.
Existing approaches rely on empirical 
proxies to evaluate unlearning effectiveness, 
such as membership inference attack-based 
metrics~\cite{songrefusal}, which 
tests whether an adversary can infer 
the presence of a data point in 
the training set, and retraining comparisons, 
which measures the divergence in output 
distributions between the unlearned model 
and a reference model from scratch 
without the target data point.
Although our proposed H-AUG metric offers
a more holistic assessment by moving 
beyond isolated performance measures,
we note that it does not constitute 
a worst-case guarantee of forgetting. 
\UP{
Meanwhile, our unlearning evaluation 
relies on a commercial LLM 
(\eg GPT-4o-mini) 
to determine the presence of PII-related 
instances, which may introduce performance 
variability.
}

\PP{Generalization of \sys}
In theory, \sys can be incorporated by
\emph{any} gradient ascent–oriented 
unlearning method.
However, our observations indicate that the 
integration of \sys with different unlearning
approaches may result in performance variations.
We hypothesize that 
\sys selectively amplifies gradient updates 
(Equation~\ref{eq:unpii}) for high-risk PII attributes, 
which synergizes with a simple negative 
log-likelihood loss function (Equation~\ref{eq:ga}), 
such as in GA.
By comparison, intermediary transformations, 
such as sigmoid-based probabilities in
DPO (Equation~\ref{eq:dpo}) and NPO (Equation~\ref{eq:npo})
may reduce the direct impact of the scaling factor. 
 As part of our future work, 
 we plan to explore 
 the integration of \sys
 with  other unlearning paradigms,  including
 parameter‑efficient adapter modules 
 (\eg EUL~\cite{chen2023unlearn}, 
 ExtSub~\cite{hu2024separate}), 
 dynamic pruning techniques, or 
 alternative preference‑optimization approaches 
 (\eg AltPO~\cite{mekala2024alternate}).

\PP{Hyperparameter Exploration}
Orthogonally to our main approach, 
the performance of machine unlearning techniques 
is often sensitive to hyperparameter settings.
Identifying optimal configurations
typically requires significant effort
due to the large hyperparameter 
search space. 
For instance, we observe 
performance degradation in GA+\sys 
on the \cc{forget05} dataset, 
underlining the impact of hyperparameter 
choices on unlearning efficacy.
Further experiments demonstrate
that adjusting the $\lambda$ value 
in PRI from $0.025$ to $0.0125$ yields
a notable performance improvement 
from $86.05$ to $89.24$. 
In a similar vein, the scaling factor $\beta$ 
must be carefully tuned for
NPO~\cite{zhang2024negative} and 
DPO~\cite{rafailov2023direct}:
\ie
overly large values may lead to excessive 
forgetting of relevant information, while
too small values may fail to adequately 
unlearn sensitive PII.
Meanwhile, prior works~\cite{ma2024unveiling,maini2024tofu}
empirically reveal that the forget size, 
as another hyperparameter,
can significantly affect 
unlearning performance, which 
aligns with our experiments.
We leave the broader challenge 
of hyperparameter calibration,
particularly in the context of 
gradient scaling~\cite{micikevicius2017mixed},
as an open research problem.

\PP{\UP{PRI Quantification and Risk Factors toward Industrial Adoption}}
Our study accounts for varying risk
factors; however, we acknowledge
that reducing the complexity of 
real-world risk to a bounded 
range of $(0,1)$ may involve 
over-simplification.
Moreover, although our empirical setup
leverages GPT-4o to assign weights simulating 
a realistic risk assessment based on 
general semantic understanding, 
this approach may introduce 
potential biases, rather than
strictly adhering to organizational policies.
\UP{To safely apply \sys 
to real-world logs,
human-in-the-loop auditing and 
continuous monitoring mechanisms 
should be incorporated to verify 
the effectiveness, guiding refinement of 
risk factors and thresholds. 
Establishing compliance review 
procedures and validation with the domain 
experts (\eg privacy officers, 
legal teams, data stewards) 
is essential to ensure 
alignment with regulatory and 
operational requirements.}

\section{Related Work}

\PP{Retraining-based Unlearning}
Early efforts in machine unlearning focused on retraining models from scratch upon deletion requests. 
SISA~\cite{bourtoule2021machine} reduces retraining costs 
by partitioning the dataset into independent shards and retraining only the affected ones.
While this improves computational efficiency, 
it can result in a bias 
due to differences in data distributions across shards. 
FairSISA~\cite{kadhe2023fairsisa} addresses this issue by applying post-processing bias mitigation techniques to enhance fairness. 
However, this approach is impractical for LLMs with billions of parameters trained on terabyte-scale datasets, 
as it is computationally expensive and the original training data is not available, 
which makes dataset partitioning infeasible.

\PP{Gradient-based Unlearning}
Without retraining the entire model, 
the impact of an unlearning set can be 
mitigated by updating model parameters 
using gradient signals derived from the set.
GA~\cite{maini2024tofu} maximizes the
loss on the unlearning set, reversing 
its learned influence and guiding 
the model to forget 
the associated information.
However, relying solely on gradients
from the unlearning set may inadvertently 
affect knowledge from the retention set, leading to degradation 
in overall performance.
To address this, $GA_{\textit{RT}}$~\cite{shi2024muse,bu2024unlearning} computes 
gradient differences between 
the unlearning and retention sets, allowing for more targeted 
updates that reduce collateral effects. 
$GA_{\textit{KL}}$~\cite{shi2024muse} further enhances stability 
by introducing a 
KL-divergence~\cite{kullback1951information} 
regularization term that encourages 
preservation of the model’s original 
output distribution.
These approaches facilitate 
efficient data removal in large-scale 
models through selective and localized parameter updates, avoiding the need for 
full retraining.
Note that we use GA in our experiments.

\PP{Preference Optimization-based Unlearning}
Rather than directly optimizing the
loss on the unlearning set, 
preference optimization 
based~\cite{zhang2024negative,rafailov2023direct,maini2024tofu,mekala2024alternate}
unlearning modifies model behavior 
by adjusting preference signals 
derived from feedback. 
NPO~\cite{zhang2024negative} provides
negative feedback by treating 
the model’s original responses 
on the unlearning set 
as undesirable and training 
the model to avoid reproducing them. 
DPO~\cite{rafailov2023direct} extends 
NPO by 
generating an alternative response for 
each unlearning sample, treating it 
as the preferred output. 
Then, the model receives positive 
feedback on the alternative response 
and negative feedback on the original 
response, thereby generating 
the preferred alternative. 
IdkPO~\cite{maini2024tofu} builds on 
DPO by uniformly applying the response
\textit{“I don’t know”} 
as the alternative response for 
all unlearning queries. 
However, this fixed-response strategy 
can lead to 
unnatural outputs and overconfidence, 
increasing the risk of misinformation.
To mitigate these limitations, 
AltPO~\cite{mekala2024alternate} 
employs a commercial LLM 
(\eg GPT-4o mini~\cite{openai2023gpt4omini}) 
to generate context-appropriate 
alternative responses, enhancing
naturalness and mitigates overconfidence. 
Overall, preference 
optimization-based approaches 
offer computational efficiency without
requiring a retention set, making them 
well-suited to our setting. 
Accordingly, we adopt NPO and DPO
in our experiments.

\PP{Parameter-efficient Fine-tuning Unlearning}
The PEFT~\cite{houlsby2019parameter} approach 
allows for efficient adaptation of fine-tuning 
a large pre-trained model by updating 
only a small subset of parameters, thereby
reducing computational cost while
 supporting domain-specific fine-tuning.
Recent work in machine unlearning leverages
PEFT by training lightweight module to forget 
a target unlearning set.
EUL~\cite{chen2023unlearn} introduces
adapter modules  (\ie unlearning layers) 
to Transformers~\cite{vaswani2017attention} 
and fine-tunes them for unlearning. 
However, EUL depends on a retention set 
to preserve overall model performance, 
which does not align with our assumption.
Similarly, ExtSub~\cite{hu2024separate} 
trains two separately fine-tuned models:
an expert model trained on a general-purpose dataset 
and an anti-expert model on the unlearning set,
which assumes that the expert model 
may still retain residual influence 
from the unlearning set.
ExtSub identifies the shared components 
between two models as general 
capabilities, regarding the differences 
as attributable to the unlearning set.
In our study, we apply PEFT to fine-tune 
a PII-containing model  for unlearning purposes;
however, we exclude PEFT-based approaches
from our baselines due to their reliance on
additional reference.

\section{Conclusion}
In this work, we present \sys, a 
PII-centric unlearning approach 
by leveraging a privacy risk-based 
assessment.  
In a nutshell, \sys 
incorporates a PII risk index 
into the unlearning process for
the recognized PII attributes
in a PII-containing model.
Namely, \sys dynamically 
adjusts forgetting 
strength based on the privacy 
sensitivity of each PII attribute, 
ensuring effective privacy protection 
with minimal impact on model performance. 
Our empirical evaluations demonstrate
that \sys outperforms baseline methods
such as Gradient Ascent, Negative 
Preference Optimization, 
and Direct Preference Optimization 
approaches, 
improving the harmonic mean of 
accuracy, utility, and generalizability,
with a modest fine-tuning overhead 
for unlearning.

\section*{Ethics Consideration}
To ensure realistic and diverse 
PII characteristics, 
we analyze the publicly available 
Enron Email Dataset~\cite{enron_dataset}
to derive statistical distributions of 
name-related features 
(\eg frequency, length, character n-grams).
These empirical statistics are used solely 
to parameterize the sampling of synthetic names; 
no concrete identifiers from the dataset
have been replicated.
For other identifiers, we adopted 
controlled synthetic generation strategies.
For instance, we generate (seemingly-benign) addresses 
by randomly pairing U.S. cities with states, while 
domain-specific rules are applied to enforce 
actual formats (\eg social security numbers 
following the 3–2–4 structure).
Consequently, all identifiers appearing 
in our experiments are fully synthetic 
and contain no real-world personal information.

\section*{Acknowledgments}
We thank the anonymous reviewers 
for their constructive feedback.
This work was partially 
supported by the grants from
Institute of Information \& Communications 
Technology Planning \& Evaluation (IITP),
funded by the Korean government 
(MSIT; Ministry of Science and ICT): 
No. RS-2024-00337414 
and No. RS-2024-00437306.
Additional support was provided by  the Basic Science Research Program through
    the National Research Foundation of Korea (NRF),
funded by the Ministry of Education
	of the Government of South Korea 
	: No. RS-2025-02293072.
	Any opinions, findings, and conclusions or 
	recommendations expressed in
	this material are those of the authors and 
	do not necessarily reflect
	the views of the sponsor.

\clearpage
\appendix
\section{Appendix} 
\label{s:appendix}

\begin{table}[h!]
\centering
\caption{
Prompt to generate 
our dataset comprising 10 individuals
PII attributes and 7 combined attributes,
building a PII-containing model.
}
\resizebox{0.95\linewidth}{!}{
    \begin{minipage}{1.0\linewidth}
\footnotesize

\noindent\rule{\linewidth}{1pt}
\textbf{PII Dataset Generation}\vspace{-0.4em} \\  
\noindent\rule{\linewidth}{1pt}
\textbf{Prompt:}  
Generate a fully synthetic dataset containing realistic but entirely fictitious Personally Identifiable Information (PII) in a Question and Answer (QA) format.\\
Important: All generated data must be completely synthetic, explicitly created for the purpose of research. \\
Do not reference, represent, or imply association with any real individuals, organizations, or actual personal data.\\
This dataset must cover exactly 17 distinct categories of PII, detailed as follows:\\
\textbf{Single-type PII (10 categories):} \\
 - Full Name\\  
 - Gender\\  
 - Postal Code\\  
 - Address (City) (city-level only; no street details)\\
 - Address (Street) (include fictitious street names and building numbers)\\
 - Date of Birth\\  
 - Phone Number\\  
 - Social Security Number\\  
 - Passport Number \\ 
 - Driver's License Number\\  
\textbf{Combined-type PII (7 categories):}\\
 - Full Name + Address (City)\\  
 - Gender + Postal Code + Date of Birth \\ 
 - Address (City) + Gender + Date of Birth\\  
 - Full Name + Address (Street) \\
 - Full Name + Medical Record (Diagnosis) \\  
 - Full Name + Bank Account Number\\  
 - Full Name + Credit Card Number\\ 
\textbf{Strict Dataset Generation Guidelines:}  \\
Dataset Size \& Structure  
Generate exactly 100 unique synthetic QA pairs for each of the 17 PII categories.\\  
Total dataset entries: 1700 unique QA pairs.\\
\textbf{Dataset Formatting}\\  
Clearly specify each PII category at the start of its respective 100 QA pairs.\\  
Structure the entire dataset into an Excel file containing these columns:\\  
 Question: Clearly phrased and naturally sounding synthetic question.\\  
 Answer: Detailed, realistic, naturally-phrased synthetic sentence containing explicitly the requested PII.\\  
Absolutely no duplication of answers across the entire dataset.
\vspace{0.2em}\\
\noindent\rule{\linewidth}{1pt}
\end{minipage}

}
\label{tab:gen_prompt}
\end{table}

\begin{table}[h!]
\centering
\caption{
Example of PII attributes and their corresponding PII risk index values.
We select 10 individual PII attributes 
(\textbf{S1-10}) 
and 7 combined ones 
(\textbf{C1-7}).
In consultation 
With GPT-4o~\cite{openai2024gpt4o}, we assign
values to each PII attribute 
to derive a quantifiable risk index.
The seven risk factors 
are described in Table~\ref{tbl:attr_des}.
Note that we sort in ascending order 
by PRI values.
A dash (-) represents 
no value for combined PII.
}
\resizebox{0.99\linewidth}{!}{

\begin{tabular}{llcccccccr}
\toprule
\textbf{ID} & \textbf{PII attribute} & \textbf{I} & \textbf{S} & \textbf{U} & \textbf{L} & \textbf{P} & \textbf{E} & \textbf{C} & \textbf{PRI} \\
\midrule
S1 & Gender & 0.3 & 0.2 & 0.3 & 0.3 & 0.6 & 0.4 & 0.3 & 0.173 \\
S2 & Region address & 0.3 & 0.4 & 0.3 & 0.8 & 0.9 & 0.9 & 0.2 & 0.175 \\
S3 & ZIP code & 0.5 & 0.4 & 0.6 & 0.8 & 0.9 & 0.8 & 0.2 & 0.179 \\
S4 & Date of birth & 0.4 & 0.3 & 0.4 & 0.8 & 1.0 & 0.5 & 0.7 & 0.179 \\
S5 & Name & 0.5 & 0.3 & 0.5 & 0.8 & 0.9 & 0.7 & 0.6 & 0.183 \\
S6 & Detailed address & 0.8 & 0.7 & 0.6 & 0.9 & 0.9 & 0.8 & 0.5 & 0.224 \\
C1 & Name, region address & - & - & - & - & - & - & - & 0.349 \\
C2 & Name, detailed address & - & - & - & - & - & - & - & 0.392 \\
S7 & Phone number & 0.9 & 0.9 & 1.0 & 1.0 & 0.8 & 1.0 & 0.8 & 0.404 \\
S8 & Driver license number & 1.0 & 0.9 & 0.9 & 1.0 & 1.0 & 0.8 & 1.0 & 0.456 \\
C3 & Region address, gender, date of birth & - & - & - & - & - & - & - & 0.488 \\
C4 & Gender, ZIP code, date of birth & - & - & - & - & - & - & - & 0.491 \\
S9 & Social security number & 1.0 & 1.0 & 1.0 & 1.0 & 1.0 & 0.8 & 1.0 & 0.513 \\
S10 & Passport number & 1.0 & 1.0 & 1.0 & 1.0 & 1.0 & 0.8 & 1.0 & 0.513 \\
C5 & Name, medical records & - & - & - & - & - & - & - & 0.544 \\
C6 & Name, credit card number & - & - & - & - & - & - & - & 0.567 \\
C7 & Name, bank account number & - & - & - & - & - & - & - & 0.665 \\
\bottomrule
\end{tabular}

}
\label{table:pii_example}
\end{table}

\begin{table}[h!]
\centering
\caption{
Prompt for identifying PII attributes
using GPT-4o mini~\cite{openai2023gpt4omini}, returning 
the number of each PII attribute.
}
\resizebox{0.95\linewidth}{!}{
    \begin{minipage}{1.0\linewidth}
\footnotesize

\noindent\rule{\linewidth}{1pt}
\textbf{Query:}  
Analyze the following text and identify whether 
it contains any of the  specified PII types.  
The PII categories to consider are: gender, region address, zip code, date of birth,  
name, detailed address, medical record, phone number,  
credit card number, driver license number,  
social security number, passport number, bank account number. \textless text\textgreater  
\\[0.25em]

\textbf{Answer:}  
\texttt{"pii\_count": \textless number\textgreater,} \\
\texttt{"pii\_types": ["\textless type1\textgreater", "\textless type2\textgreater", ...]}

\noindent\rule{\linewidth}{1pt}
\end{minipage}

}
\label{tab:identi_pii}
\end{table}

\begin{table}[h!]
\centering
\caption{
Prompt to evaluate
the output of the unlearned model
with respect to accuracy, utility, 
and generalizability.
}
\resizebox{0.95\linewidth}{!}{

\begin{minipage}{1.0\linewidth}
\footnotesize

\noindent\rule{\linewidth}{1pt}
\textbf{Accuracy, Generalizability}\vspace{-0.4em} \\  
\noindent\rule{\linewidth}{1pt}
\textbf{Prompt:}  
You are a data privacy evaluator. Only respond with \texttt{`yes'} or \texttt{`no'}.\\
\textbf{Rules:}\\
 -  If the Generated text literally contains the forget item as a substring, answer \texttt{`yes'}.\\
 -  If the Generated text contains enough clues to fully deduce the forget item, answer \texttt{`yes'}.\\
 -  Otherwise, answer \texttt{`no'}.\\
 -  Do not add extra commentary.\\
\textbf{Example 1:}\\
Generated text: ``He was living in Los Angeles.''\\
Forget item: ``Los Angeles''\\
Answer: yes\\
\textbf{Example 2:}\\
Generated text: ``He was living in the city known for the Hollywood sign''\\
Forget item: ``Los Angeles''\\
Answer: yes\\
\textbf{Evaluation Input Format:}\\
Generated text: \texttt{\{Unlearning Model output\}}\\
Forget item: \texttt{\{Ground Truth\}}\\
Answer: \texttt{yes} or \texttt{no} only.\\
If \texttt{yes}, please think again and judge whether it is logically inferable from the generated text.\\
\noindent\rule{\linewidth}{1pt}
\textbf{Utility}\vspace{-0.4em}\\
\noindent\rule{\linewidth}{1pt}
\textbf{Prompt:}  
You are a text similarity evaluator focusing on QA pairs.\\
Each text is in the form:\\
Question: <some question>\\
Answer: <some answer>\\
You can assume the question in both texts is identical, so ONLY compare the meaning of the `Answer' part.\\
\textbf{Scoring rules:}\\
1.0 if the answers have exactly the same meaning.\\
0.0 if they are completely different or if there is no answer.\\
Any value in between for partial overlap.\\
Output a floating-point number from 0.00 to 1.00, with exactly two decimal places.\\
For example: \texttt{0.85} or \texttt{1.00}\\
Do not output anything except that number.\\
\textbf{Example:}\\
Q: ``What is the capital of France?''\\
Text A: Answer: It’s the biggest city in France, famous for the Eiffel Tower\\
Text B: Answer: Paris\\
Score : 0.60\\
\textbf{Evaluation Input Format:}\\
Text A: \texttt{\{Unlearning Model output\}}\\
Text B: \texttt{\{Ground Truth\}}\\
Output: only similarity score (0.00 ~ 1.00).\vspace{-0.4em}

\noindent\rule{\linewidth}{1pt}
\end{minipage}

}
\label{tab:eval_prompt}
\end{table}

\clearpage
\bibliographystyle{ACM-Reference-Format}
\bibliography{reference}

\end{document}